\documentclass[10pt,twocolumn,letterpaper]{article}

\usepackage[pagenumbers]{cvpr} 

\usepackage{graphicx}
\usepackage{amsmath}
\usepackage{bbm}
\usepackage{amssymb}
\usepackage{booktabs}
\usepackage{enumitem}
\usepackage{makecell}

\usepackage[pagebackref,breaklinks,colorlinks]{hyperref}

\usepackage{multirow}
\usepackage[table,xcdraw]{xcolor}

\usepackage[export]{adjustbox}
\usepackage{graphicx}

\usepackage{algorithm}
\usepackage{algorithmic}

\usepackage[capitalize]{cleveref}
\crefname{section}{Sec.}{Secs.}
\Crefname{section}{Section}{Sections}
\Crefname{table}{Table}{Tables}
\crefname{table}{Tab.}{Tabs.}

\def\cvprPaperID{482} 
\def\confName{CVPR}
\def\confYear{2023}

\begin{document}

\title{Open-World Multi-Task Control Through\\ Goal-Aware Representation Learning and Adaptive Horizon Prediction}


\author{
    \textbf{Shaofei Cai}$^{1,2}$, \textbf{Zihao Wang}$^{1,2}$, \textbf{Xiaojian Ma}$^{3}$, \textbf{Anji Liu}$^{3}$, \textbf{Yitao Liang}$^{1,4}$ \\
    \multicolumn{1}{c}{\textbf{Team CraftJarvis}} \\
    $^{1}$Institute for Artificial Intelligence, Peking University\\
    $^{2}$School of Intelligence Science and Technology, Peking University \\
    $^{3}$Computer Science Department, University of California, Los Angeles \\
    $^{4}$Beijing Institute for General Artificial Intelligence (BIGAI)\\
    {\tt \small \{caishaofei,zhwang\}@stu.pku.edu.cn,xiaojian.ma@ucla.edu} \\
    {\tt \small liuanji@cs.ucla.edu,yitaol@pku.edu.cn}
}

\maketitle

\begin{abstract}

We study the problem of learning goal-conditioned policies in Minecraft, a popular, widely accessible yet challenging open-ended environment for developing human-level multi-task agents. We first identify two main challenges of learning such policies: 1) the indistinguishability of tasks from the state distribution, due to the vast scene diversity,
 and 2) the non-stationary nature of environment dynamics caused by partial observability. To tackle the first challenge, we propose Goal-Sensitive Backbone (GSB) for the policy to encourage the emergence of goal-relevant visual state representations. To tackle the second challenge, the policy is further fueled by an adaptive horizon prediction module that helps alleviate the learning uncertainty brought by the non-stationary dynamics. Experiments on 20 Minecraft tasks show that our method significantly outperforms the best baseline so far; in many of them, we double the performance. Our ablation and exploratory studies then explain how our approach beat the counterparts and also unveil the surprising bonus of zero-shot generalization to new scenes (biomes). We hope our agent could help shed some light on learning goal-conditioned, multi-task agents in challenging, open-ended environments like Minecraft. The code is released at \url{https://github.com/CraftJarvis/MC-Controller}.

\end{abstract}


\section{Introduction}
\label{sec:intro}

\begin{figure}[t]
    \centering
    \includegraphics[scale = 0.37]{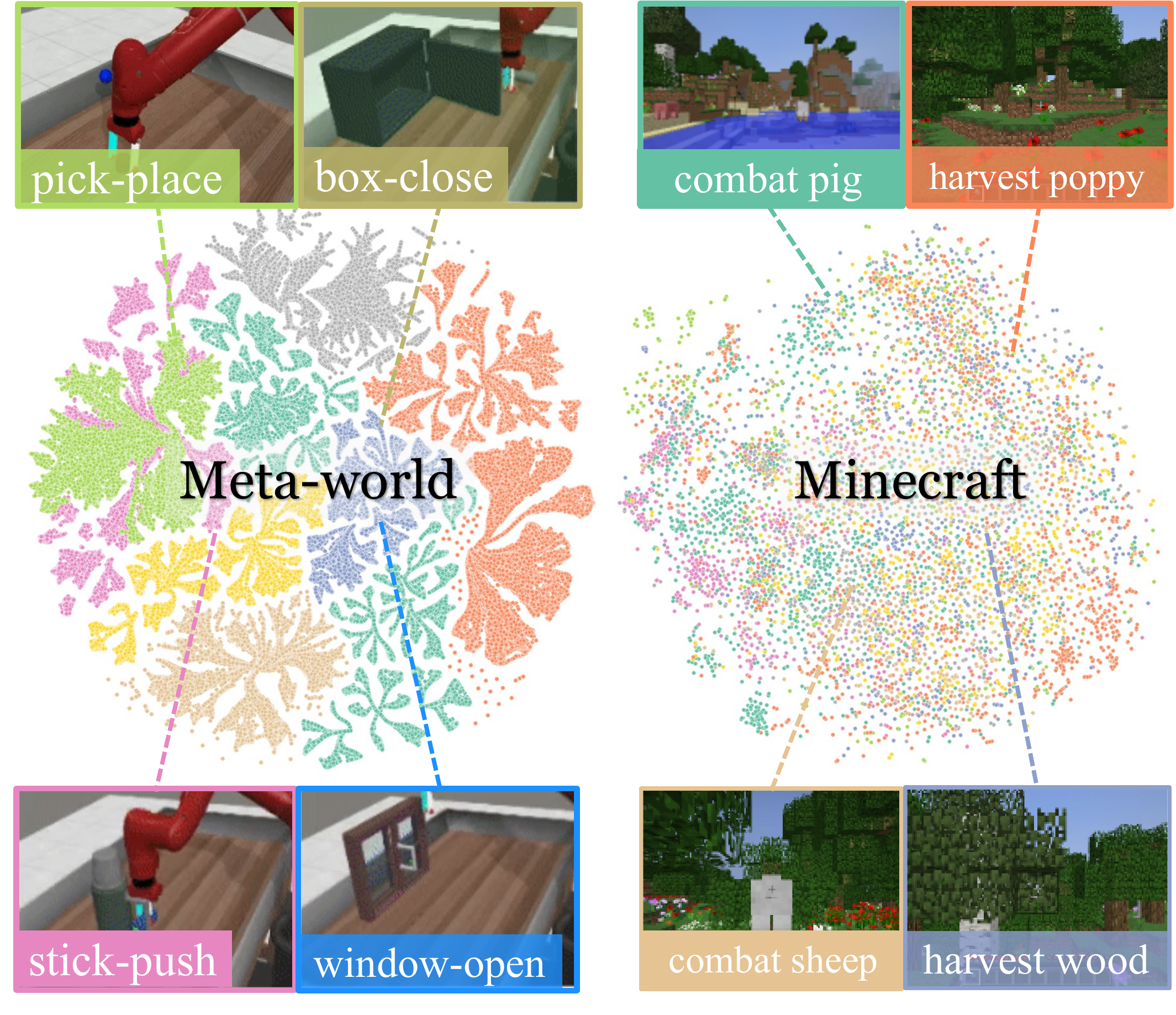}
    \vskip -0.1in
    \caption{
        Comparison of states between Meta-world~\cite{metaworld}~(left) and Minecraft~\cite{malmo}~(right) based on t-SNE visualization. 
        The points with the same color represent states from the trajectories that complete the same task. It can be seen that the states are much more distinguishable in terms of tasks in Meta-world than in Minecraft, implying the higher diversity of states and tasks in open worlds like Minecraft over traditional multi-task agent learning environments like Meta-world.  
    }
    \label{tsne}
    \vskip -0.2in
\end{figure}
Building agents that can accomplish a vast and diverse suite of tasks in an open-ended world is considered a key challenge towards devising generally capable artificial intelligence~\cite{gato,flamingo,saycan,gpt3}. In recent years, environments like Minecraft have drawn much attention from the related research communities~\cite{minerl,minedojo,minerl-rel-1,minerl-rel-2,minerl-rel-3}, since they are not only popular, and widely accessible, but also offer an open-ended universe with myriad of tasks, making them great platforms for developing human-level multi-task agents. 
Although groundbreaking successes have been observed in many challenging sequential decision-making problems such as Atari\!~\cite{atari}, Go\!~\cite{go}, and MOBA games\!~\cite{starcraft, starcraft2, starcraft3}, such successes have not been transferred to those open worlds. 
To understand the gap and design corresponding solutions, we need to first understand the distinct challenges brought by these environments. Let's take Minecraft~\cite{malmo} as an example: there are over twenty types of landscapes ranging from flat lands like Savannah and desert to rough mountains with forests and caves. These diverse landscapes also enable countless tasks that could be achieved by the agents: mining, harvesting, farming, combating, constructing, etc. Compared to canonical agent learning environments like Go~\cite{go}, Atari~\cite{atari}, and robotic control suite~\cite{meta-world,mujoco,dm_control}, Minecraft provides a substantially more diverse distribution of states thanks to the rich scenes and tasks built with the game, making it exceptionally difficult to extract the pivotal task-relevant visual state representations for goal-conditioned policies. To help our readers understand the significance of this challenge, we visualize the states from trajectories that complete some tasks in Minecraft and Meta-world~\cite{meta-world} (a popular multi-task learning environment but with fewer states and tasks) in Fig.~\ref{tsne}. States of different tasks are annotated with different colors. Clearly, the states in Minecraft are much less distinguishable in terms of tasks than in Meta-world. Therefore goal-conditioned policies are more likely to struggle in mapping those states and tasks (served as goals) to actions.

Another grand challenge in an open-ended environment like Minecraft hails from the setting of such games, where an agent can only have very limited observations of the world. For example, in MineDoJo~\cite{minedojo} (a recent agent benchmark built on Minecraft), the observation space comprises a first-person view image and a list of possessed items. However, many more aspects of the surroundings remain hidden from the agents. That is, the agent now has to work with a \textbf{partially observable environment}. A plague embedded with such an environment is \textit{non-stationary dynamics}, which makes it almost impossible to predict what will happen next. Therefore, the distances from states to the current goal become much less clear due to the world uncertainty, leading to less distinguishable states in terms of goal completeness and more faulty decisions emitted by the goal-conditioned policies.

This paper aims at mitigating both aforementioned challenges that emerge from most open-world environments. First, we observe that the architecture of the policy network is crucial to learning goal-relevant visual state representations that allow goal-conditioned actions in domains with low inter-goal state diversity (cf. Fig.~\ref{tsne}). To this end, we propose Goal-Sensitive Backbone (GSB), which enables effective learning goal-conditioned policies over 20 tasks in the Minecraft domain. Next, to mitigate the challenge posed by the partially observed and non-stationary environment, we introduce horizon as an extra condition for the policy and a corresponding horizon prediction module. Specifically, the policy is also \textit{explicitly} conditioned on the remaining time steps till achieving certain goals (i.e., distance-to-goal). We find it significantly boosts the performance of our agents in open-world multi-task domains. However, the ground-truth distance-to-goal is unavailable during evaluation. To fix this problem, we train a horizon prediction module and feed the estimated distance-to-goal to the horizon commanding policy in evaluation. 
This leads to a $27\%$ gain in average success rate under the multi-task settings. 

We evaluate the proposed approaches based on the simple yet effective behavior cloning algorithm~\cite{gcil}. 
The experiments are conducted in three common biomes. 
In multi-task settings, our proposed method outperforms the baseline in terms of success rate and precision by a large margin. 
It also achieves consistent improvement in single-task settings. Our ablation and exploratory studies then explain how our approach beat the counterparts and also unveil the surprising bonus of zero-shot generalization to new scenes (biomes).

To summarize, targeting two identified challenges distinct to open worlds, our contributions are threefold:
\setlength{\leftmargini}{0.85em}
\begin{itemize}[topsep=0.8pt]
 \setlength\itemsep{0.6pt}
\item We propose Goal-Sensitive Backbone (GSB), a neural network that enables effective learning goal-relevant visual state representations at multiple levels for goal-conditioned policies, aiming at addressing the challenge of diverse state distribution in open-ended environments.
\item We further introduce adaptive horizon prediction to explicitly condition the policy on the distance from the current state to the goal, yielding much better performances in a partially observable open-ended environment with non-stationary dynamics.
\item We conduct extensive studies on the popular yet challenging Minecraft domain with baselines and our proposed method. The results demonstrate superior advantages of our approach over the counterparts in terms of both success rate and precision of task completion. 
\end{itemize}

\section{Preliminaries}

\noindent \textbf{Goal-conditioned policy}, as its name suggests, is a type of agent's policy $\pi$ for decision-making that is conditioned on goals besides states. Specifically, we denote $\pi(a|s, g)$ as a goal-conditioned policy that maps the current state $s$ and goal $g$ to an action $a$. Compared to the canonical formulation of policy where the goal is absent, the goal-conditioned policy offers flexibility of learning \textit{multi-task} agent as it allows different behaviors for different tasks by simply altering the goal. There are multiple ways to specify the goal, e.g., natural language instructions~\cite{saycan} and goal images~\cite{habitat}. 

\noindent \textbf{Goal-conditioned imitation learning} is a simple yet effective way to learn goal-conditioned policies. Specifically, $\pi(a|s,g)$ is optimized by imitating the demonstrations $\mathcal{D}$, where $\mathcal{D} = \{\tau^1, \tau^2, \tau^3, \dots\}$ is a collection of trajectories $\tau^i$. A trajectory is a sequence of states, actions, and goals, defined as $\tau^i = \{(s^i_t, a^i_t, g^i)\}^{T}_{t=0}$, where $T$ is the trajectory length. 
The imitation learning objective is to maximize the likelihood of the action in demonstrations when attempting to reach the desired goal
\begin{equation}
J_{IL}(\pi) = \mathbb{E}_{\tau \sim \mathcal{D}} \big[ \sum\nolimits_{t=0}^{T} \text{log} \ \pi(a_t | s_t, g) \big]. 
\end{equation}

\noindent \textbf{Notation. } 
At each timestep, our architecture takes in a tuple $(\boldsymbol{s}_t, \boldsymbol{a}_t, h_t, \boldsymbol{g}, \boldsymbol{a}_{t-1})$ as the input, where $\boldsymbol{s}_t = \{\boldsymbol{o}^{I}_{t}, \boldsymbol{o}^{E}_{t}\}$, $\boldsymbol{o}^{I}_{t}$ is the raw image observation, $\boldsymbol{o}^{E}_{t}$ is the extra observation provides by the environments. 
$h_t$ comes from the demonstration. 
$\tilde{h}_t$ and $\tilde{\boldsymbol{a}}_t$ are the predicted horizon and action, respectively. 
For simplicity, we also use the same symbols ($\boldsymbol{o}^{E}_{t}, \boldsymbol{g}, \boldsymbol{a}_{t-1}$) to represent their embeddings.

\begin{figure}[t]
    \centering
    \includegraphics[scale = 0.42]{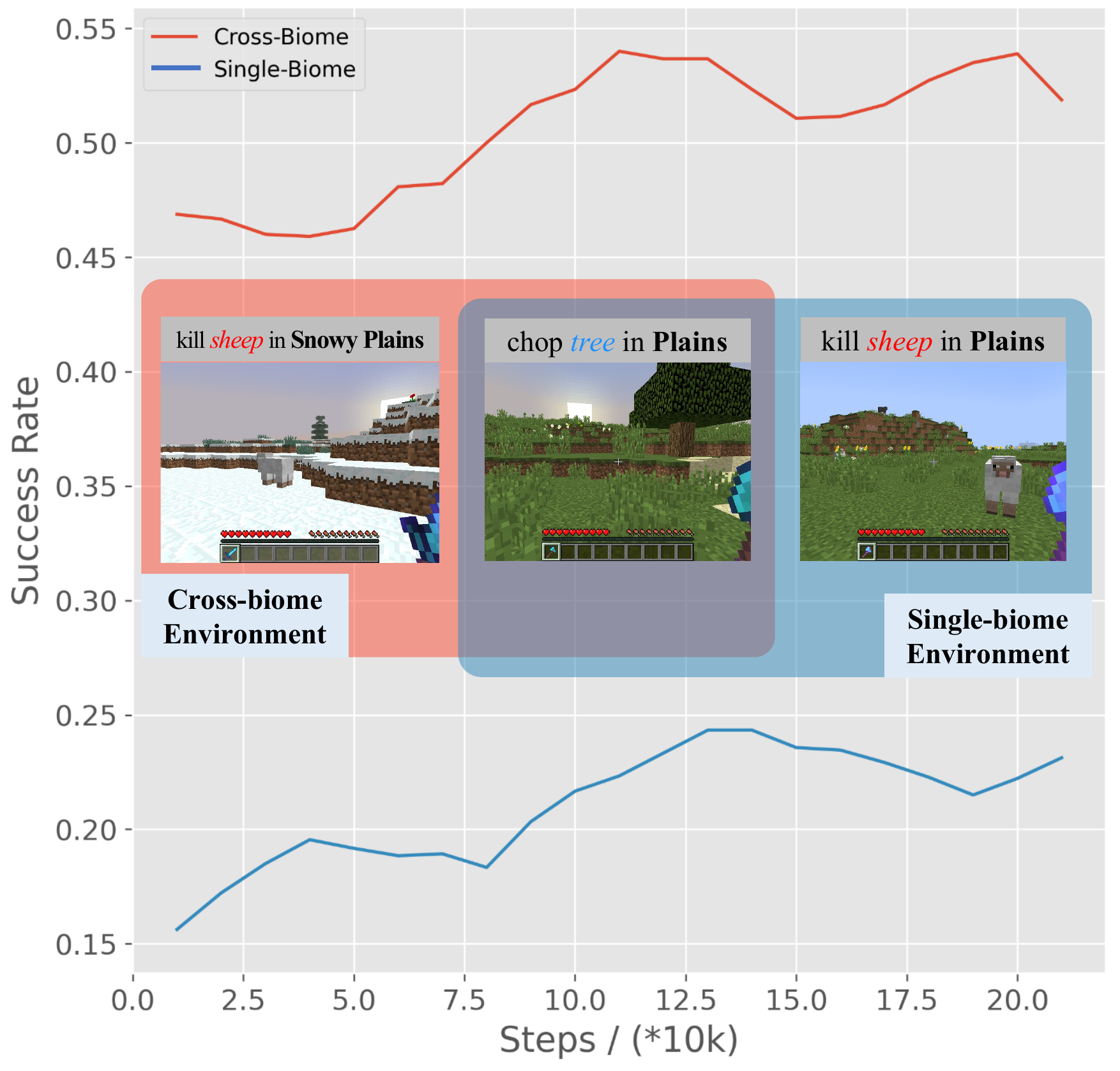}
    
    \caption{
        Demonstrations of the cross-biome environment and the more challenging single-biome environment. The challenge comes from the fact that the agent needs to learn diverse behaviors in similar states conditioned on different goals. 
    }
    \label{fig:demons}
    \vskip -0.2in
\end{figure}

\section{Method}

In this section, we describe the proposed algorithm for learning goal-conditioned policies that are capable of completing various preliminary tasks in open-world domains. First, we revisit and provide a detailed illustration of the identified challenges in open-world domains (\S\ref{sec:challenges}). Aiming at solving these challenges, we proceed to introduce the proposed goal-sensitive backbone (\S\ref{sec:backbone}) and adaptive horizon prediction module (\S\ref{sec:horizon}). Finally, we provide an overview of the proposed method in Section~\ref{sec:overview}.


\subsection{Challenges}
\label{sec:challenges}

As demonstrated in Section~\ref{sec:intro}, the \textbf{first} major challenge of open-world environments is the indistinguishability of states in terms of different goals (cf. Fig.~\ref{tsne}). That is, it is often hard to identify the task/goal by looking at individual states. Compared to environments with clear goal indicators in their states, agents in open-world domains need to learn goal-conditioned diverse behaviors under similar states.

This challenge can be reflected by the illustrative experiment in Fig.~\ref{fig:demons}. 
Two multi-task environments are created based on the Minecraft domain. Both environments consist of two preliminary tasks: collect logs and hunt sheep, where the former can be done by chopping trees and the latter requires the agent to slaughter sheep. Both tasks require the agent to first locate and approach the corresponding target. As shown in Fig.~\ref{fig:demons} (center), in the single-biome environment (blue blob in Fig.~\ref{fig:demons}), the agent is tasked to collect logs and hunt sheep both inside a randomly generated plain area with grass, trees, and various mobs. In contrast, in the cross-biome environment~(red blob in Fig.~\ref{fig:demons}), whenever the agent is tasked to hunt sheep, it is spawned randomly in a snowy plain. Although different in visual appearance, snowy plains and plains have very similar terrains, so the difficulty of each task in the cross-biome environment is similar to its counterpart in the single-biome environment. The main consequence of this change is that the agent can determine its goal by solely looking at the current state, which mimics the setting of Meta-World in Fig.~\ref{tsne}(left).


We collect demonstrations by filtering successful trajectories played by VPT~\cite{VPT} (see \S\ref{para:collection} for more details) and use behavior cloning to train multi-task policies on both environments. Perhaps surprisingly, as shown in Fig.~\ref{fig:demons}, despite the minor difference, performance in the single-biome environment is significantly weaker than in the cross-biome one. This clearly demonstrates that the common practice of directly concatenating observation features and goal features 
suffer from learning diverse actions (e.g., locate trees, find sheep) given similar observations. 
In contrast, in the cross-biome environment,  the difficulty of the two tasks fundamentally remains the same, yet the agent only needs to learn a consistent behavior in each biome (i.e., plains and snow fields). This alleviates the need to learn goal-conditioned diverse behaviors in similar states and leads to a better success rate.

The \textbf{second} key challenge comes from the partial observability of the game and non-stationary environment dynamics. Specifically, in Minecraft, the biome and mobs surrounding the agent are generated procedurally and randomly after each reset. Further, only a small fraction of the whole terrain is visible to the agent in one observation, leading to more uncertainty of the world. 
From the perspective of learning goal-conditioned policies, the distances from states to the current goal will become much less clear compared to canonical learning environments like Atari~\cite{goexplore}. We refer to Appendix~\ref{sec:horizon_distribuition} for more discussion on this. Since the goal-conditioned policies also rely on distinguishable states in terms of goal completeness, they're more likely to make wrong decisions as a result of world uncertainty.

\begin{figure*}[t]
    \centering
    \includegraphics[scale = 0.50]{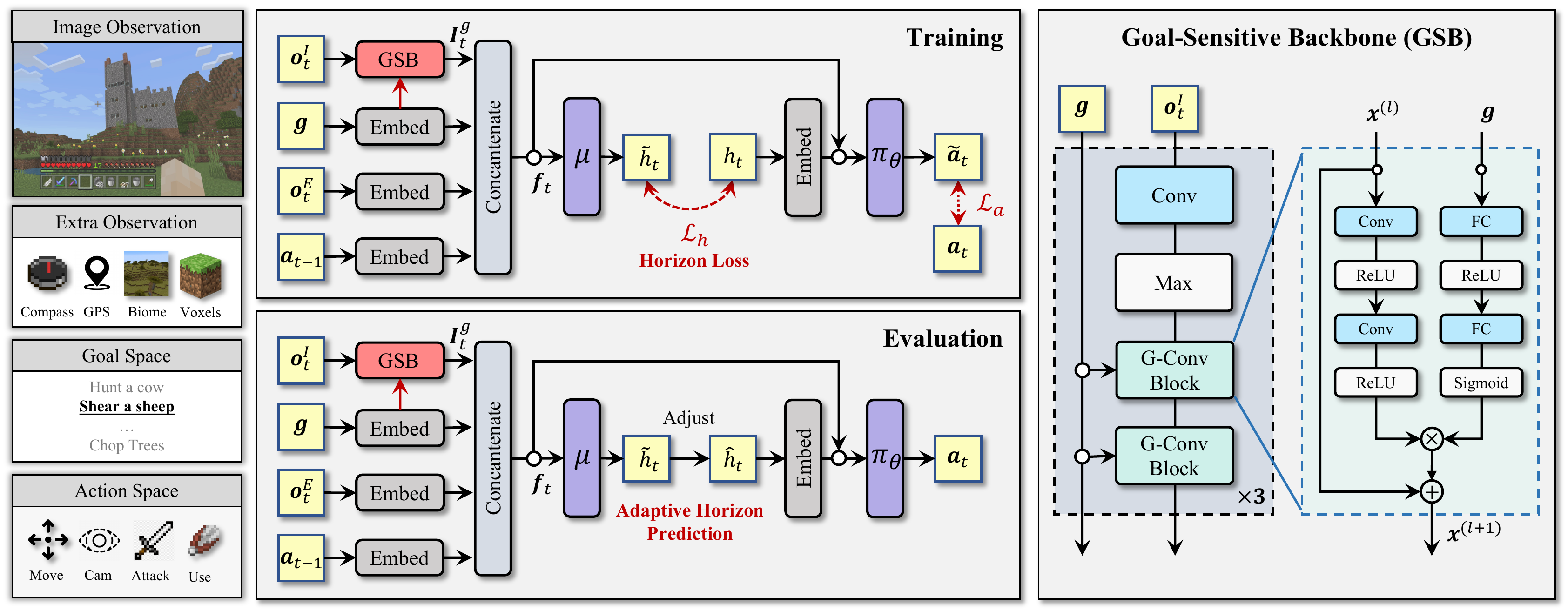}
    
    \caption{
        \textbf{Our Goal-conditioned Policy Architecture}. Our contributions are in red and purple.
        \textbf{Right:} The \emph{goal-sensitive backbone} (GSB) is a key component to incentivize goal-condition behaviors. It consists of a stack of g-conv blocks. It takes the image observation $\boldsymbol{o}^{I}_{t}$ and the goal embedding $\boldsymbol{g}$ as input, and outputs the goal-attended visual representation $\boldsymbol{I}^{g}_{t}$. 
        The multimodal joint representation $\boldsymbol{f}_t$ is the concatenation of visual representation $\boldsymbol{I}^{g}_{t}$, goal embedding $\boldsymbol{g}$, extra observation embedding $\boldsymbol{o}^E_t$ and previous action embedding $\boldsymbol{a}_{t-1}$. 
        The horizon prediction module $\mu$ uses it to predict the horizon $\tilde{h}_t$ while 
        the horizon commanding policy $\pi_\theta$ uses it to predict the action $\tilde{\boldsymbol{a}}_t$. 
        \textbf{Top:} During the training, the predicted horizon $\tilde{h}_t$ is only used to compute the horizon loss $\mathcal{L}_h$. 
        The policy is conditioned on $h_t$ that comes from the demonstration. 
        \textbf{Bottom:} During the evaluation, the policy is conditioned on the predicted horizon $\tilde{h}_t$ which needs to be adjusted.
    }
    \label{pipeline}
    \vskip -0.2in
\end{figure*}

\subsection{Incentivize Goal-Conditioned Behavior with Stacked Goal-Sensitive Backbone}
\label{sec:backbone}
As elaborated in Section~\ref{sec:challenges}, learning goal-conditioned policies becomes extremely hard when states collected from trajectories that accomplish different tasks are indistinguishable. 
While certain algorithmic design choices could improve multi-task performance in such open-world environments, we find that the structure of the policy network is a key factor towards higher episode reward.
Specifically, we observe that existing CNN-based backbones can excel at completing many single tasks (e.g., hunt cow, collect stone), but struggle to learn goal-conditioned behavior when training on the tasks in a goal-conditioned manner. This motivates the need to properly fuse goal information into the network. Despite the existence of various feature fusion approaches such as concatenation and Bilinear layers \cite{bilinear}, they all perform poorly even with a moderate number of tasks. 
This motivates the need to carry goal information into multiple layers of the network.
Specifically, we propose goal-sensitive backbone (GSB), which effectively blends goal information to the state features at multiple levels. 
As shown in Fig.~\ref{pipeline} (right), GSB is composed with multiple goal convolution blocks (g-conv block), which are obtained by augmenting the vanilla convolution block with a goal branch. 
Functionally, it can provide deep feature fusion between multi-level visual features and the goal information. As we will proceed to show in Section~\ref{sec:ablation}, adding GSB can lead to significant performance boost in multi-task environments.
The g-conv block processes its input visual features $\boldsymbol{x}^{(l)} \in \mathbb{R}^{C \times H \times W}$ with two convolution layers
\begin{equation}
    \hat{\boldsymbol{x}}^{(l)} = \text{ReLU}(\text{Conv}(\text{ReLU}(\text{Conv}(\boldsymbol{x}^{(l)})))). 
\end{equation}
Meanwhile, it maps the goal embedding $\boldsymbol{g}$ to the same feature space as the intermediate features $\hat{\boldsymbol{x}}^{(l)}$ with two fully-connected layers, decribed as 
\begin{equation}
\hat{\boldsymbol{g}}^{(l)} = \text{FC}(\text{ReLU}(\text{FC}(\boldsymbol{g}))). 
\end{equation}
The goal feature $\hat{\boldsymbol{g}}^{(l)}$ is then used to modulate the intermediate features $\hat{\boldsymbol{x}}^{(l)}$ channel-wise. 
By adding a residual connection \cite{resnet}, the output feature $\boldsymbol{x}^{(l+1)}$ is expressed by 
\begin{equation}
    \boldsymbol{x}^{(l+1)} = \sigma(\hat{\boldsymbol{g}}^{(l)}) \odot \hat{\boldsymbol{x}}^{(l)} + \boldsymbol{x}^{(l)},
\end{equation}
where $\sigma(\cdot)$ is the sigmoid function and $\odot$ is the element-wise product. 
This channel-wise modulation encourages the module to focus on goal-specific regions and discard the background information by adaptively weighing the channel importance. 
We highlight that the g-conv block can be plugged into any convolution backbone to improve its capability of extracting goal-aware visual features. 
The proposed goal-sensitive backbone is constructed by replacing 6 convolution blocks of the widely-adopted Impala CNN~\cite{impala} to g-conv blocks. 
In our experiments, a GSB is used to compute goal-conditioned state features $\boldsymbol{I}^{g}_t=\text{GSB}(\boldsymbol{o}_{t}^{I}, \boldsymbol{g})$.
Such an idea of fusing condition information into the backbone layer by layer was also used by some prior works~\cite{film1, film2, film3, film4}. Here, we demonstrate that it works in a critical role for open-world multi-task control.




\subsection{Combat World Uncertainty with Adaptive Horizon Prediction}
\label{sec:horizon}

To address the challenge brought by the uncertainty of the world, we need to ensure the goal-conditioned policies to be more aware of goal-completeness given the current state. We observe that conditioning the policy additionally on the number of remaining steps toward achieving a goal, i.e., distance-to-goal, or \textbf{horizon}, can significantly improve the accuracy of predicted actions on held-out offline datasets \cite{udrl, gcsl}. 
Here, we define the horizon $h_t := T - t$, where $T$ is the trajectory length, as the remaining time steps to complete the given goal. 
This motivates the design of a horizon commanding policy 
$\pi_{\theta}: \mathcal{S} \times \mathcal{G} \times \mathcal{H} \rightarrow \mathcal{A}$ that takes a state $s$, a goal $g$, and a horizon $h$ as inputs and outputs an action $a$. 
A key problem of the horizon commanding policy is that it cannot be directly used for evaluation: during gameplay, horizon is unknown as it requires completing the whole trajectory. To fix this problem, we introduce an additional horizon prediction module, which estimates the horizon given a state $s$ and a goal $g$. Combining the two modules together, we can apply the fruitful horizon commanding policy during gameplay.

Both modules can be trained efficiently with dense supervision. 
Specifically, the horizon commanding policy $\pi_{\theta}$ can be learned by any policy loss specified by RL algorithms. 
For example, when behavior cloning is used, $\pi_{\theta}$ can be optimized by minimizing the loss
\begin{equation}
    \mathcal{L}_{a} = -\text{log} \  \pi_\theta(\boldsymbol{a}_t|h_t, \boldsymbol{f}_t),
\end{equation}
where $\boldsymbol{f}_t$ is the joint representation of the state and goal embedded by a neural network (see \S\ref{sec:overview}). 
The horizon prediction module is trained by a supervised learning loss 
\begin{equation}
    \mathcal{L}_{h} = -\text{log} \ \mu(h_t|\boldsymbol{f}_t),
\end{equation}
\noindent where $\mu$ is a network that predicts the horizon.

During the evaluation, after computing the embedding $\boldsymbol{f}_t$ for $s_t$ and $g$, the horizon prediction module $\mu$ is first invoked to compute an estimated horizon $\tilde{h}_t = \mu(\boldsymbol{f}_t)$. This predicted horizon can then be fed to the horizon commanding policy to compute the action distribution $\pi_{\theta} (\boldsymbol{a}_t | \tilde{h}_t, \boldsymbol{f}_t)$.
In practice, we observe that feeding an adaptive version of $\tilde{h}_t$, defined as $\hat{h}_t := \max (\tilde{h}_t - c, 0)$ ($c$ is a hyperparameter), to $\pi_{\theta}$ leads to better performance. We hypothesize that this advantageous behavior comes from the fact that by supplying the adaptive horizon $\hat{h}_t$, the agent is encouraged to choose actions that lead to speedy completion of the goal. The effectiveness of the adaptive horizon will be demonstrated in Section~\ref{sec:ablation}.


\subsection{Model Summary}
\label{sec:overview}

As shown in Fig.~\ref{pipeline}, our model sequentially connects the proposed goal-sensitive backbone, horizon prediction module, and horizon commanding policy. At each time step $t$, the image observation and goal information are first fed forward into the goal-sensitive backbone to compute goal-aware visual feature $\boldsymbol{I}^{g}_t$.
The visual feature is then fused with additional input information including the extra observation embedding $\boldsymbol{o}^{E}_t$, the goal embedding $\boldsymbol{g}$, and the previous action embedding $\boldsymbol{a}_{t-1}$ by concatenation and a feed-forward network:
\begin{equation}
    \boldsymbol{f}_t = \text{FFN}(\big[\boldsymbol{I}^{g}_t \parallel \boldsymbol{o}^{E}_t \parallel \boldsymbol{g} \parallel  \boldsymbol{a}_{t-1}    \big]).
\end{equation}
Then, $\boldsymbol{f}_t$ is input to the horizon prediction module to predict horizon $\tilde{h}_t = \mu(\boldsymbol{f}_t)$. 
And the horizon commanding policy takes in the horizon and features $\boldsymbol{f}_t$ to compute the action. 
When trained with behavior cloning, the overall objective function is $\mathcal{L} = \mathcal{L}_a + \mathcal{L}_h$. During the evaluation, the adaptive horizon $\hat{h}_t$ is fed to the horizon commanding policy in replacement of $\tilde{h}_t$.

\section{Experiments}

This section analyzes and evaluates the proposed goal-sensitive backbone and the adaptive horizon prediction module in the open-world domain Minecraft. To minimize performance variation caused by the design choices in RL algorithms, we build the proposed method on top of the simple yet effective behavior cloning algorithm. In Section~\ref{sec:experimental_setup}, we first introduce three suites of tasks; the agent is asked to collect and combat various target objects/mobs with indistinguishable states conditioned on different goals (challenge \#1) and non-stationary environment dynamics (challenge \#2). 
Single-task and multi-task performance on the benchmarks is evaluated and analyzed in Section~\ref{sec:results}, and ablation studies are conducted in Section~\ref{sec:ablation}. Finally, we unveil the surprising bonus of zero-shot generalization to new scenes and tasks in Section~\ref{sec:generation}.



\subsection{Experimental Setup}
\label{sec:experimental_setup}

\begin{figure}[t]
    \centering
    \includegraphics[scale = 0.50]{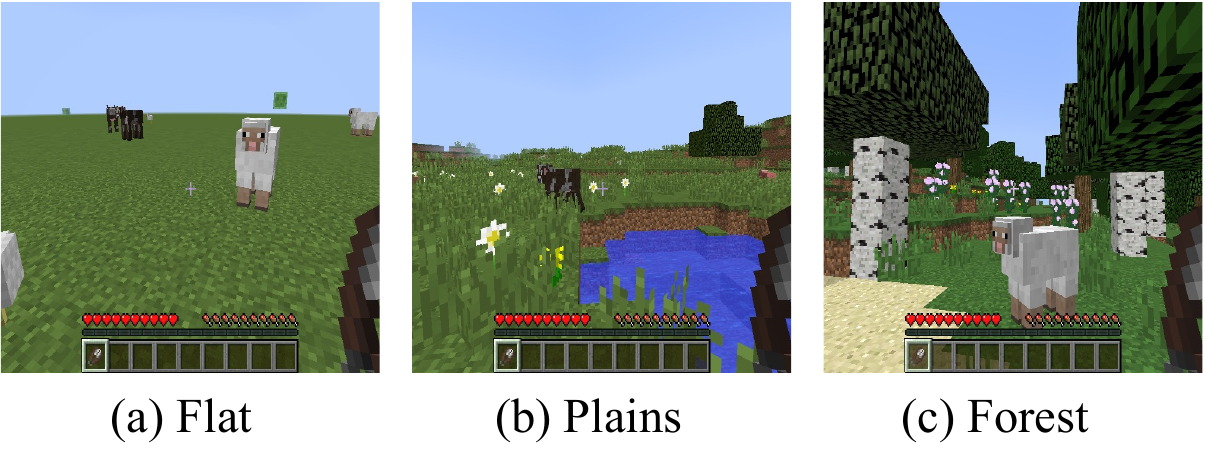}
        \vskip -0.2in
    \caption{
        Snapshots of the RGB camera view in three biomes. 
    }
    \label{fig:env}
    \vskip -0.2in
\end{figure}


\noindent \textbf{Environment and task.}
To best expose the challenges described in Sections~\ref{sec:intro} and \ref{sec:challenges}, a key design principle of our benchmark environments is to task the agent to complete multiple preliminary tasks in similar yet highly randomized scenes. By specifying the biome that surrounds the agent, Minecraft provides a perfect way to create such environments. Specifically, as shown in Fig.~\ref{fig:env}, every biome has unique and consistent observations; randomness comes from the fact that the terrain is generated randomly in each episode. To evaluate the scalability of the proposed method in terms of the number of tasks, we choose \textbf{\texttt{Plains}} and \textbf{\texttt{Forest}}, the two most common biomes that contain a large number of resources and mobs. 

In addition to the two challenges, \textbf{\texttt{Plains}} and \textbf{\texttt{Forest}} also add unique difficulties to learning goal-conditioned policies. Specifically, although we have better views in \textbf{\texttt{Plains}}, the resources/targets are located further away from the agent and require more exploration. In contrast, there exist more occlusions and obstacles in \textbf{\texttt{Forest}}.

The \textbf{\texttt{Plains}} benchmark consists of four tasks: harvest {\texttt{oak wood}}~(\includegraphics[scale=0.15,valign=c]{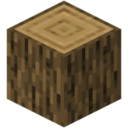}), and Combat {\texttt{sheep}}~(\includegraphics[scale=0.15,valign=c]{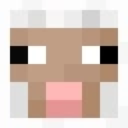}), {\texttt{cow}}~(\includegraphics[scale=0.15,valign=c]{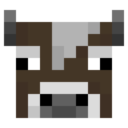}), {\texttt{pig}}~(\includegraphics[scale=0.15,valign=c]{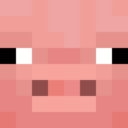}).
In the \textbf{\texttt{Forest}} benchmark, the agent is tasked to complete thirteen tasks: combat {\texttt{sheep}}~(\includegraphics[scale=0.15,valign=c]{minecraft/sheep}), {\texttt{cow}}~(\includegraphics[scale=0.15,valign=c]{minecraft/cow}), {\texttt{pig}}~(\includegraphics[scale=0.15,valign=c]{minecraft/pig}), harvest {\texttt{dirt}}~(\includegraphics[scale=0.15,valign=c]{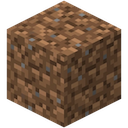}), {\texttt{sand}}~(\includegraphics[scale=0.15,valign=c]{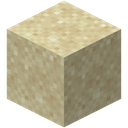}), {\texttt{oak wood}}~(\includegraphics[scale=0.15,valign=c]{minecraft/oak_wood}), {\texttt{birch wood}}~(\includegraphics[scale=0.15,valign=c]{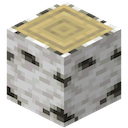}), {\texttt{oak leaves}}~(\includegraphics[scale=0.15,valign=c]{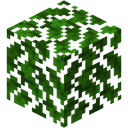}), {\texttt{birch leaves}}~(\includegraphics[scale=0.15,valign=c]{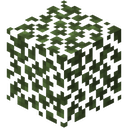}), {\texttt{wool}}~(\includegraphics[scale=0.15,valign=c]{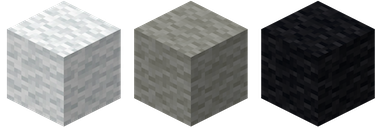}), {\texttt{grass}}~(\includegraphics[scale=0.15,valign=c]{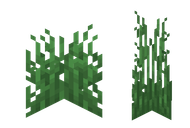}), {\texttt{poppy}}~(\includegraphics[scale=0.15,valign=c]{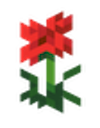}), {\texttt{orange tulip}}~(\includegraphics[scale=0.15,valign=c]{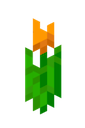}).

In addition to the above two benchmarks, we also test the agent on a ``hunt animals'' benchmark based on the \textbf{\texttt{Flat}} biome, which contains a flattened world. Specifically, the agent needs to combat {\texttt{sheep}}~(\includegraphics[scale=0.15,valign=c]{minecraft/sheep}), {\texttt{cow}}~(\includegraphics[scale=0.15,valign=c]{minecraft/cow}), {\texttt{pig}}~(\includegraphics[scale=0.15,valign=c]{minecraft/pig}), {\texttt{spider}}~(\includegraphics[scale=0.15,valign=c]{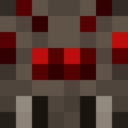}), {\texttt{polar bear}}~(\includegraphics[scale=0.15,valign=c]{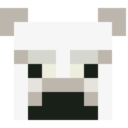}), {\texttt{chicken}}~(\includegraphics[scale=0.15,valign=c]{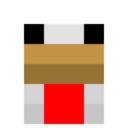}), {\texttt{donkey}}~(\includegraphics[scale=0.15,valign=c]{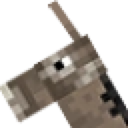}), {\texttt{horse}}~(\includegraphics[scale=0.15,valign=c]{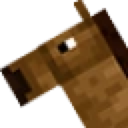}), {\texttt{wolf}}~(\includegraphics[scale=0.15,valign=c]{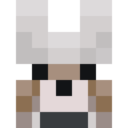}), {\texttt{llama}}~(\includegraphics[scale=0.15,valign=c]{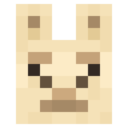}), {\texttt{mushroom cow}}~(\includegraphics[scale=0.15,valign=c]{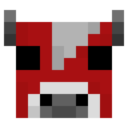}) in the \textbf{\texttt{Flat}} environment. Compared to other benchmarks, the challenge of \textbf{\texttt{Flat}} comes from the fact that the mobs are constantly wondering around, which makes it hard to locate and approach the correct target.

We adopt the original observation space provided by MineDoJo\cite{minedojo}, which includes a RGB camera-view, yaw/pitch angle, GPS location, and the type of $3 \times 3$ blocks surrounding the agent. We discretize the original multi-discrete action space provided by MineDojo into 42 discrete actions. Details are included in Appendix~\ref{sec:space}.


\noindent \textbf{Data collection pipeline.}
\label{para:collection}
One significant downside of behavior cloning algorithms is the need for high-quality and densely-labeled trajectories, which often requires enormous human effort to collect. To mitigate this problem, we collect goal-conditioned demonstrations by filtering successful trajectories from gameplays by pretrained non-goal-conditioned policies. Specifically, we adopt Video Pre-Training (VPT) \cite{VPT}, which is trained on tremendous amount of non-goal-conditioned gameplays. We rollout the VPT policy in the three benchmarks and record all episodes that accomplishes any of the defined goals. These trajectories are then converted to a goal-conditioned demonstration dataset. Please refer to Appendix~\ref{sec:data_collection_pipeline} for detailed settings and efficiency analysis of our data collection pipeline.

\noindent \textbf{Evaluation.} 
During the evaluation, the maximum episode length is set to 600, 600, and 300 on the \textbf{\texttt{Flat}}, \textbf{\texttt{Plains}} and \textbf{\texttt{Forest}} benchmarks, respectively. \textbf{\texttt{Plains}} and \textbf{\texttt{Forest}} are given more time steps since, in these environments, the agent needs more time to locate and approach the target.
We use \textit{Success Rate} and \textit{Precision} as our evaluation metrics. A gameplay is successful if the agent completes the goal within the episode. 
Precision is defined as the number of times the specified goal is achieved divided by the total number of goals completed in an episode. It measures how well the agent can be aware of the specified goal, instead of simply accomplishing any goal during gameplay.


\subsection{Experimental Results}\label{sec:results}

We first focus on the simpler single-task learning setting in order to isolate the challenge introduced by non-stationary dynamics and partial observability (\S\ref{sec:single_task}). We then examine whether the proposed method can better address both challenges by examining its multi-task performance (\S\ref{sec:multi_task}).


\subsubsection{Single task experiments}\label{sec:single_task}

We select three typical tasks, i.e., harvest log, hunt cow, and hunt sheep, from the \textbf{\texttt{Plains}} benchmark for single-task training. We compare the proposed method against the following baselines. First, MineAgent \cite{minedojo} is an online RL algorithm that leverages pretrained state representations and dense reward functions to boost training. BC~({\footnotesize{VPT}}) \cite{VPT}, BC~({\footnotesize{CLIP}}) \cite{minedojo}, and BC~({\footnotesize{I-CNN}}) \cite{impala} are variants of the behavior cloning algorithm that use different backbone models (indicated in the corresponding brackets) for state feature extraction. The backbones are finetuned with the BC loss (see Appendix~\ref{sec:implementation} for more details). 

Results are reported in Table~\ref{tab:single}. First, we observe that even the individual tasks are extremely challenging for online RL algorithms such as MineAgent, even its networks are pretrained on Minecraft data. 
We attribute this failure to its inconsistent dense reward when facing a hard-exploration task (e.g., the additional provided reward is not consistently  higher when the agent is moving closer to a target object).
Next, compared to BC~({\footnotesize{I-CNN}}) that uses a randomly initialized impala CNN model, the Minecraft-pretrained backbones in BC~({\footnotesize{VPT}}) and BC~({\footnotesize{CLIP}}) do not bring any benefit. This could be caused by the lack of plasticity, i.e., the ability to learn in these well-trained models, echoing similar findings in computer vision and RL \cite{dohare2021continual}. Finally, our approach outperforms all baseline methods, especially in terms of precision. This demonstrates that our method is more robust against non-stationary dynamics and partially observable observations.

\definecolor{Gray8}{gray}{0.8}
\definecolor{Gray9}{gray}{0.9}
\begin{table}[t]
\footnotesize
\caption{Results of \textbf{single-goal} tasks (\S\ref{sec:single_task}) on \textbf{\texttt{Plains}}. }
\vskip -0.1in
\setlength{\tabcolsep}{0.8 mm}
\label{tab:single}
\resizebox{\linewidth}{!}{
\renewcommand\arraystretch{1.0}
\begin{tabular}{@{}ccccccc@{}}
\toprule
\multirow{2}{*}[-0.2em]{Method} & \multicolumn{3}{c}{\textbf{Success Rate (\%)}} & \multicolumn{3}{c}{\textbf{Precision (\%)}} \\ \cmidrule(l){2-4} \cmidrule(l){5-7}   
  & \includegraphics[scale=0.2,valign=c]{minecraft/oak_wood} & \includegraphics[scale=0.2,valign=c]{minecraft/cow} & 
 \includegraphics[scale=0.2,valign=c]{minecraft/sheep} & \includegraphics[scale=0.2,valign=c]{minecraft/oak_wood} & \includegraphics[scale=0.2,valign=c]{minecraft/cow} & \includegraphics[scale=0.2,valign=c]{minecraft/sheep} \\ \midrule
MineAgent\cite{minedojo} & $00_{\pm00}$ & $01_{\pm00}$ & $01_{\pm00}$ & \makecell[c]{--} & \makecell[c]{--} & \makecell[c]{--}\\
BC~({\scriptsize{CLIP}})\cite{minedojo} & $18_{\pm06}$ & $26_{\pm05}$ & $25_{\pm06}$ & $51_{\pm08}$ & $43_{\pm08}$ & $44_{\pm05}$ \\
BC~({\scriptsize{VPT}})\cite{VPT} & $22_{\pm08}$ & $27_{\pm06}$ & $22_{\pm06}$ & $58_{\pm09}$ & $46_{\pm05}$ & $42_{\pm05}$ \\
\rowcolor{Gray9} BC~({\scriptsize{I-CNN}})\cite{impala} & $45_{\pm05}$  & $46_{\pm04}$ & $48_{\pm07}$ & $\boldsymbol{86}_{\pm05}$ & $55_{\pm12}$ & $45_{\pm07}$ \\ \midrule
\rowcolor{Gray8} \textbf{\texttt{Ours}} & $\boldsymbol{50}_{\pm07}$ &  $\boldsymbol{58}_{\pm10}$ &  $\boldsymbol{60}_{\pm08}$ &  $83_{\pm10}$ &  $\boldsymbol{75}_{\pm10}$ &  $\boldsymbol{75}_{\pm06}$ \\ \bottomrule
\end{tabular}
}
\end{table}

\subsubsection{Multi-task experiments }\label{sec:multi_task}

We move on to evaluate the proposed method on the three multi-task benchmarks introduced in Section~\ref{sec:experimental_setup}. 
The baseline includes three behavior cloning methods (we use ``MT-BC'' as an abbreviation of multi-task behavior cloning). 
We also include two variations of our method: one without the goal-sensitive backbone, and the other without the adaptive horizon prediction module.
Results on the \textbf{\texttt{Plains}}, \textbf{\texttt{Flat}}, and \textbf{\texttt{Forest}} environments are reported in Table~\ref{tab:syn}, respectively. 
First, we observe that our method significantly outperforms all baselines in terms of both success rate and precision in all three benchmarks. Moreover, scaling up the number of tasks does not necessarily deteriorate the performance of our method. Specifically, we compare the average success rate on the \textbf{\texttt{Plains}} and \textbf{\texttt{Flat}} benchmark, which contain 4 and 9 tasks, respectively. While the baselines struggle to maintain their success rate on the \textbf{\texttt{Flat}} environment, our approach is capable of maintaining high performance despite the increased number of tasks. Putting together, results on multi-task benchmarks clearly demonstrate the superiority of our method when facing open-world environments with the two elaborated challenges (cf. \S\ref{sec:challenges}).

\definecolor{Gray8}{gray}{0.8}
\definecolor{Gray9}{gray}{0.9}

\begin{table}[t]
  \footnotesize
  \setlength{\tabcolsep}{0.3 mm}
  \caption{Results of \textbf{multi-goal} tasks (\S\ref{sec:multi_task}) on three biomes. }
  \vskip -0.1in
  \label{tab:syn}
  \resizebox{\linewidth}{!}{
  \renewcommand\arraystretch{1.1}
  \begin{tabular}{@{}ccccccc@{}}
  \toprule
  \multirow{2}{*}[-0.2em]{\textbf{Method}} & \multicolumn{3}{c}{\textbf{Avg. Success Rate (\%)}} & \multicolumn{3}{c}{\textbf{Avg. Precision (\%)}} \\ \cmidrule(l){2-4} \cmidrule(l){5-7}   
    & Plains & Flat & Forest & Plains & Flat & Forest \\ \midrule
  MT-BC~({\scriptsize{VPT}})\cite{VPT} & $25_{\pm06}$ & $17_{\pm05}$ & $15_{\pm04}$ & $22_{\pm05}$ & $17_{\pm03}$ & $14_{\pm04}$ \\
  MT-BC~({\scriptsize{CLIP}})\cite{minedojo} & $22_{\pm05}$ & $14_{\pm03}$ & $14_{\pm03}$ & $23_{\pm04}$ & $15_{\pm03}$ & $13_{\pm03}$ \\
  MT-BC~({\scriptsize{I-CNN}})\cite{impala} & $25_{\pm02}$ & $18_{\pm02}$ & $15_{\pm03}$ & $23_{\pm04}$ & $14_{\pm02}$ & $13_{\pm03}$ \\ \midrule
  \rowcolor{Gray9} MT-BC~({\scriptsize{w/ GSB}}) & $32_{\pm05}$ & $36_{\pm03}$ & $19_{\pm05}$ & $43_{\pm06}$ & $36_{\pm02}$ & $17_{\pm03}$ \\
  \rowcolor{Gray9} \textbf{\texttt{Ours}}~({\scriptsize{I-CNN}}) & $31_{\pm06}$ & $31_{\pm04}$ & $18_{\pm02}$ & $22_{\pm03}$ & $28_{\pm04}$ & $15_{\pm04}$ \\
  \rowcolor{Gray8} \textbf{\texttt{Ours}}~({\scriptsize{w/ GSB}}) & $\boldsymbol{55}_{\pm09}$ & $\boldsymbol{57}_{\pm09}$ & $\boldsymbol{30}_{\pm06}$ & $\boldsymbol{70}_{\pm09}$ & $\boldsymbol{50}_{\pm06}$ & $\boldsymbol{29}_{\pm06}$ \\ \bottomrule
  \end{tabular}
  }
  \vskip -0.2in
  \end{table}


\subsection{Ablation Study}\label{sec:ablation}

\noindent \textbf{Ablation study on goal-sensitive backbone.}\label{sec:ablation_backbone}
To examine the effectiveness of our proposed goal-sensitive backbone, we compare the following two groups of architectures: 1) \textbf{\texttt{Ours}}~({\scriptsize{I-CNN}}) v.s. \textbf{\texttt{Ours}}~({\scriptsize{w/ GSB}}), 2) MT-BC~({\scriptsize{I-CNN}}) v.s. MT-BC~({\scriptsize{w/ GSB}}). 
The key distinction between the groups is whether the backbone employs a standard Impala CNN or a goal-sensitive backbone.
As depicted in Table~\ref{tab:syn}, our findings indicate that the goal-sensitive backbone consistently enhances performance in terms of both success rate and precision across all environments. 
Remarkably, in the \textbf{\texttt{Flat}} biome, our approach with the goal-sensitive backbone attains a $26\%$ and $22\%$ performance improvement in success rate and precision, respectively. 
This demonstrates that the goal-sensitive backbone effectively fuses the goal information into visual features and leads to goal-aware behavior.

\begin{figure}[t]
    \centering
    \includegraphics[scale = 0.25]{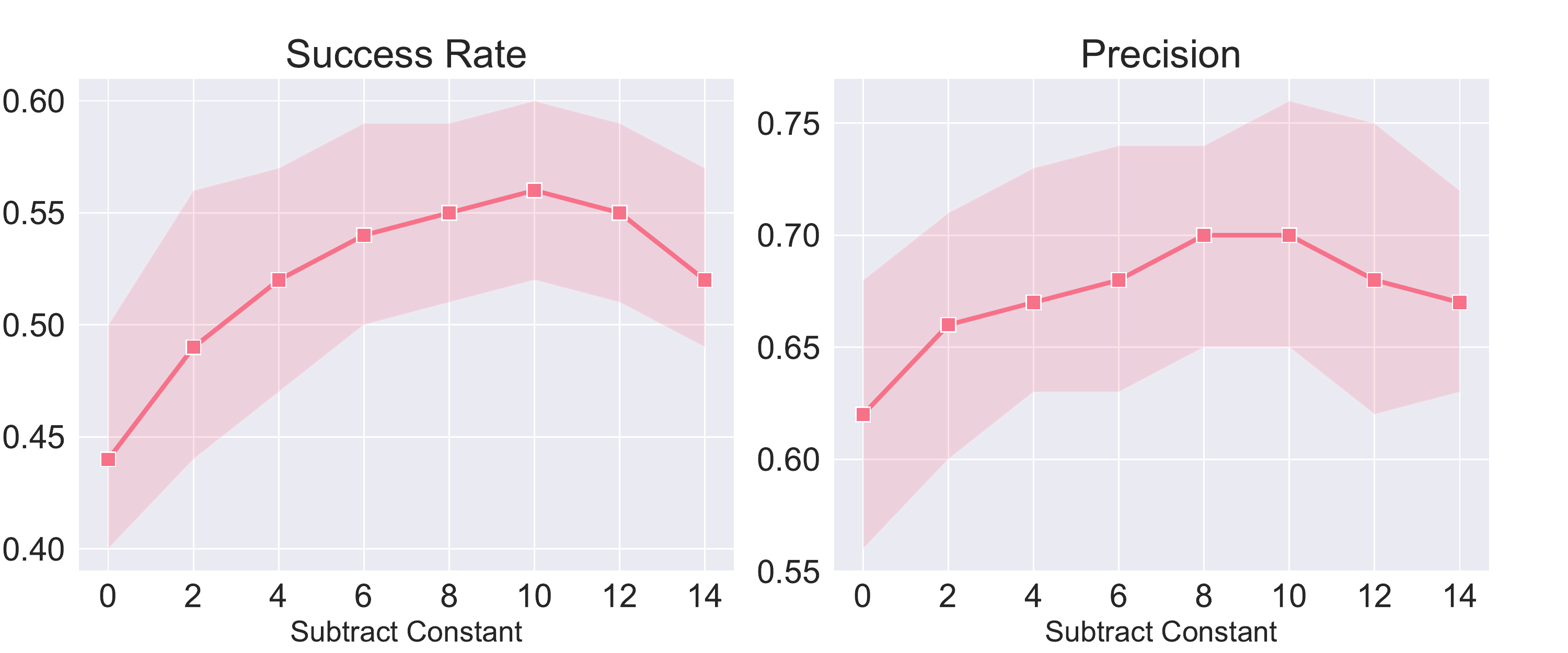}
        \vskip -0.1in
    \caption{
        Multi-task performance as a function of subtracting the horizon constant $c$. Results show that setting $c$ to a small constant lead to better overall performance as it incentivizes the agent to exhibit behaviors that lead to faster task completion.
    }
    \label{fig:ablation}
    \vskip -0.2in
\end{figure}

\noindent \textbf{Parameter sensitivity on horizon prediction.}\label{sec:ablation_horizon}
To investigate the sensitivity of the horizon-based control policy to the constant $c$ (outlined in \S\ref{sec:horizon}), we perform experiments with $c$ values ranging from 0 to 14. We train and evaluate the model using the multi-task setting on the $\textbf{\texttt{Flat}}$ benchmark, shown in Figure~\ref{fig:ablation}. Our findings indicate that within the 0 to 10 range, decreasing $c$ enhances performance, while further reduction leads to decline. This implies that subtracting a small constant from the predicted horizon-to-goal yields a more effective policy. However, subtracting a larger value results in performance deterioration, as attaining the goal within such a limited horizon may be unfeasible. 


\definecolor{Gray9}{gray}{0.9}
\definecolor{Gray8}{gray}{0.8}
\definecolor{Gray5}{gray}{0.5}
\definecolor{Gray7}{gray}{0.7}

\begin{table}[t]
  \footnotesize
  \caption{Additional ablation experiments on \textbf{\texttt{Plains}} biome. }
  \vskip -0.1in
  \setlength{\tabcolsep}{0.8 mm}
  \resizebox{\linewidth}{!}{
  \renewcommand\arraystretch{1.0}
  \label{ablation}
  \begin{tabular}{@{}llcc@{}}
  \toprule
  \textbf{\#} & \textbf{Method}  & \textbf{Avg. SR (\%)}  & \textbf{Avg. P (\%)}      \\ \midrule 
  \rowcolor{Gray9} 1  & Ours (GSB + horizon pred)                      & $\boldsymbol{55}_{\pm09}$ & $\boldsymbol{70}_{\pm09}$ \\ 
  \rowcolor{Gray8} 2  & Ours + RNN                     & $\boldsymbol{65}_{\pm07}$ & $\boldsymbol{67}_{\pm08}$ \\ 
  3  & Ours \textminus~horizon pred + RNN & $39_{\pm08}$ & $51_{\pm08}$ \\ 
  4  & Ours \textminus~horizon pred          & $35_{\pm08}$ & $45_{\pm15}$ \\
  \midrule
  5  & w/o horizon loss & $47_{\pm06}$ & $54_{\pm08}$ \\ 
  6  & w/o extra obs             & $50_{\pm07}$ & $69_{\pm07}$ \\ 
  7  & w/o language condition    & $25_{\pm03}$ & $26_{\pm05}$ \\
  \bottomrule
  \end{tabular}
  }
  \vskip -0.2in
  \end{table}

\begin{table}[]
  \footnotesize
    \caption{The success rate (SR) under condition-free policy. }
    \vskip -0.1in
    \setlength{\tabcolsep}{1.0 mm}
    \renewcommand\arraystretch{1.0}
    \resizebox{\linewidth}{!}{
    \label{conditionfree}
    \begin{tabular}{@{}cccccc@{}}
    \toprule
    \textbf{Goal}  & \includegraphics[scale=0.16,valign=c]{minecraft/oak_wood} & \includegraphics[scale=0.16,valign=c]{minecraft/sheep} & \includegraphics[scale=0.16,valign=c]{minecraft/cow} & \includegraphics[scale=0.16,valign=c]{minecraft/pig} & Avg. \\ \midrule
    \textbf{Success Rate (\%)} & $44_{\pm19}$ & $24_{\pm06}$ & $23_{\pm11}$ & $11_{\pm07}$ & $25_{\pm03}$ \\ \bottomrule
    \end{tabular}
    }
    \vskip -0.2in
\end{table}

\noindent \textbf{Comparision with recurrent architecture. }
We built two recurrent variants ( ``Ours + RNN'', ``Ours \textminus~horizon pred + RNN'') by using a GRU module to fuse the joint representation $f_t$ and optionally also removing the horizon prediction module. 
During training, the batch size, frame number, and skipping frame are set to 8, 16, and 5, respectively. 
Table \ref{ablation} (exp1 \vs exp3) shows that ``Ours \textminus~horizon pred + RNN'' becomes significantly worse, likely due to the partial observability issue ($-26\%$ SR). 
However, when combining RNN and horizon module (exp2), the performance gains significantly more than our original method ($+10\%$ SR). 
To sum up, while RNNs can aid in addressing partial observability, our findings indicate that in our open-world scenario, they are considerably more effective when combined with our horizon prediction module.

\noindent \textbf{Ablation on horizon loss, extra observation, and language condition. } 
Table \ref{ablation} demonstrates that excluding horizon loss (exp5) and extra observation (exp6) can result in a decrease of success rate by $8\%$ and $5\%$, respectively.
Furthermore, as depicted in Table \ref{conditionfree}, when the language condition is removed from the input (exp7), the policy primarily accomplishes the ``chopping tree'' task ($44\%$ SR) while scarcely completing the ``hunting pig'' task ($11\%$ SR). The tasks ``hunting sheep'' and ``hunting cow'' are executed fairly evenly (around $24\%$ SR). This is likely due to trees appearing more frequently than animals in the environment. 

\subsection{Generalization Performance}\label{sec:generation}


In the open-ended Minecraft environment, which features a variety of biomes with distinct appearances, a decent agent should be capable of generalizing across these diverse biomes. To evaluate the agent's zero-shot generalization ability in a new biome, we initially train the agent using data exclusively from the Plains biome. Subsequently, we test it in the Flat biome, where it faces the challenge of combatting {\texttt{sheep}}, {\texttt{cows}}, and {\texttt{pigs}}. Complicating the task, numerous distracting mobs, such as {\texttt{wolves}} and {\texttt{mushroom cows}}, appear in the testing biome but not in the training biome. 
The results are presented in Table~\ref{tab:cross_biome}. Our zero-shot agent demonstrates success rates comparable to those of an agent trained directly on the Flat biome. The high precision of our zero-shot agent also indicates its robust performance, even amidst numerous novel distracting mobs in the new testing biome. Therefore, we believe that our agent displays a degree of zero-shot generalization to new environments, achieved through goal-aware representation learning and adaptive horizon prediction.

\section{Related Works}
\noindent \textbf{Open-ended Environments. }
A variety of environments have been developed for open-ended agent training, such as grid worlds~\cite{open-grid-1,open-grid-2}, maze worlds~\cite{open-maze-1, open-maze-2, open-maze-3}, and indoor worlds~\cite{open-indoor-1, open-indoor-2, open-indoor-3, open-indoor-4}. Although these benchmarks have advanced agent development, they generally lack complexity in perception and task domains. This paper concentrates on Minecraft, a voxel-based 3D, first-person, open-world game centered around survival and creation. 
Microsoft introduced the first Gym-style API platform called Malmo~\cite{malmo} for Minecraft, which has spawned numerous secondary development variants. Building on Malmo, MineRL~\cite{minerl} offers a human-interface simulator and a dataset of human play demonstrations for the annual Diamond Challenge at NeurIPS~\cite{minerl-rel-1, minerl-rel-2, minerl-rel-3}. MineDoJo~\cite{minedojo}, an extension of MineRL, broadens the APIs for customizing tasks and provides thousands of pre-defined compositional tasks aimed at developing a generally capable embodied agent, which we use to evaluate our method.

\noindent \textbf{Embodied Agents in Minecraft. } 
Some prior studies have utilized a hierarchical reinforcement learning framework to develop sophisticated embodied agents. For instance, SEIHAI~\cite{SEIHAI} divides a long-horizon task into several subtasks, training an appropriate agent for each subtask and designing a scheduler to manage the execution of these agents. Similarly, JueWu-MC~\cite{juewu} adopts this concept but enhances the agent with action-aware representation learning capabilities. 
In recent times, the internet-scale pretraining paradigm has made a significant impact on embodied research in open-ended environments. VPT~\cite{VPT}, for example, undergoes pretraining on an extensive collection of online gameplay videos using imitation learning. However, it lacks the ability to process any command input. MineAgent~\cite{minedojo} takes a different approach by pretraining a language-conditioned reward function using online video-transcript pairs, which is then utilized to support multi-task reinforcement learning. 

\noindent \textbf{Progress Monitor. } 
The horizon-to-goal prediction technology has already been employed as a progress monitor in the Vision-Language Navigation (VLN) communities~\cite{progress1, progress3, progress4}. This technology aids in understanding the task structure and expediting the training procedure. Generally, current progress monitors primarily function as supplementary objectives. Their estimated progress is utilized to reassess actions or execute beam search. In contrast, our estimated horizon is explicitly incorporated into the policy network to guide agent behaviors. During inference, the horizon input can be adjusted for enhanced performance.


\begin{table}[t]
\caption{Quantitive results on generalization to a novel biome. }
\vskip -0.1in
\label{tab:cross_biome}
\resizebox{\linewidth}{!}{
\setlength{\tabcolsep}{1.4 mm}
\renewcommand\arraystretch{0.8}
\begin{tabular}{@{}p{0.5cm}cccccccc@{}}
\toprule
\multirow{2}{*}[-0.2em]{\hspace{1.0em} \textbf{Train~$\rightarrow$~Eval}} & \multicolumn{4}{c}{\textbf{Success Rate (\%)}} & \multicolumn{4}{c}{\textbf{Precision (\%)}} \\ \cmidrule(l){2-5}  \cmidrule(l){6-9} 
 &  \includegraphics[scale=0.18,valign=c]{minecraft/sheep} & \includegraphics[scale=0.18,valign=c]{minecraft/cow} & \includegraphics[scale=0.18,valign=c]{minecraft/pig} & Avg. & \includegraphics[scale=0.18,valign=c]{minecraft/sheep} & \includegraphics[scale=0.18,valign=c]{minecraft/cow} & \includegraphics[scale=0.18,valign=c]{minecraft/pig} & Avg. \\ \midrule
\multicolumn{1}{r}{\texttt{\textbf{Flat}}$\rightarrow$\texttt{\textbf{Flat}}} & $72$ & $60$ & $57$ & $\boldsymbol{63}$ & $44$ & $48$ & $54$ & $\boldsymbol{49}$\\
\multicolumn{1}{r}{\texttt{\textbf{Plains}}$\rightarrow$\texttt{\textbf{Flat}}} & $67$ & $47$ & $60$ & $\boldsymbol{58}$ & $89$ & $89$ & $70$ & $\boldsymbol{83}$ \\ \bottomrule
\end{tabular}
}
\vskip -0.2in
\end{table}

\section{Conclusion}
In this paper, we explore the issue of learning goal-oriented policies in open-world environments. We pinpoint two major challenges unique to such settings: 1) the difficulty in distinguishing tasks from the state distribution due to immense scene variety, and 2) the non-stationary nature of environmental dynamics resulting from partial observability.  We propose a goal-sensitive backbone and an adaptive horizon prediction module to overcome both. Our experiments on challenging Minecraft confirm the advantages of our proposed methods over baselines in terms of both success rate and precision of task completeness. 

\noindent \textbf{Acknowledgement.} This work was supported by the National Key R\&D Program of China 2022ZD0160301, and in part by the NSF grants \#IIS-1943641, \#IIS-1956441, \#CCF-1837129, Samsung, CISCO, and a Sloan Fellowship. We thank Hongming Xu for his engineering support.

{\small
\bibliographystyle{ieee_fullname}
\bibliography{egbib}

\begin{thebibliography}{10}\itemsep=-1pt

\bibitem{open-indoor-1}
Josh Abramson, Arun Ahuja, Arthur Brussee, Federico Carnevale, Mary Cassin,
  Stephen~R. Clark, Andrew Dudzik, Petko Georgiev, Aurelia Guy, Tim Harley,
  Felix Hill, Alden Hung, Zachary Kenton, Jessica Landon, Timothy~P. Lillicrap,
  Kory~Wallace Mathewson, Alistair Muldal, Adam Santoro, Nikolay Savinov,
  Vikrant Varma, Greg Wayne, Nathaniel Wong, Chen Yan, and Rui Zhu.
\newblock Imitating interactive intelligence.
\newblock {\em arXiv: Learning}, 2020.

\bibitem{saycan}
Michael Ahn, Anthony Brohan, Noah Brown, Yevgen Chebotar, Omar Cortes, Byron
  David, Chelsea Finn, Keerthana Gopalakrishnan, Karol Hausman, Alex Herzog,
  et~al.
\newblock Do as i can, not as i say: Grounding language in robotic affordances.
\newblock {\em arXiv preprint arXiv:2204.01691}, 2022.

\bibitem{flamingo}
Jean-Baptiste Alayrac, Jeff Donahue, Pauline Luc, Antoine Miech, Iain Barr,
  Yana Hasson, Karel Lenc, Arthur Mensch, Katie Millican, Malcolm Reynolds,
  et~al.
\newblock Flamingo: a visual language model for few-shot learning.
\newblock {\em arXiv preprint arXiv:2204.14198}, 2022.

\bibitem{VPT}
Bowen Baker, Ilge Akkaya, Peter Zhokhov, Joost Huizinga, Jie Tang, Adrien
  Ecoffet, Brandon Houghton, Raul Sampedro, and Jeff Clune.
\newblock Video pretraining {(VPT):} learning to act by watching unlabeled
  online videos.
\newblock {\em CoRR}, abs/2206.11795, 2022.

\bibitem{film2}
Anthony Brohan, Noah Brown, Justice Carbajal, Yevgen Chebotar, Joseph Dabis,
  Chelsea Finn, Keerthana Gopalakrishnan, Karol Hausman, Alex Herzog, Jasmine
  Hsu, et~al.
\newblock Rt-1: Robotics transformer for real-world control at scale.
\newblock {\em arXiv preprint arXiv:2212.06817}, 2022.

\bibitem{gpt3}
Tom Brown, Benjamin Mann, Nick Ryder, Melanie Subbiah, Jared~D Kaplan, Prafulla
  Dhariwal, Arvind Neelakantan, Pranav Shyam, Girish Sastry, Amanda Askell,
  et~al.
\newblock Language models are few-shot learners.
\newblock {\em Advances in neural information processing systems},
  33:1877--1901, 2020.

\bibitem{bias1}
Han Cai, Chuang Gan, Ligeng Zhu, and Song Han.
\newblock Tinytl: Reduce memory, not parameters for efficient on-device
  learning.
\newblock In H. Larochelle, M. Ranzato, R. Hadsell, M.F. Balcan, and H. Lin,
  editors, {\em Advances in Neural Information Processing Systems}, volume~33,
  pages 11285--11297. Curran Associates, Inc., 2020.

\bibitem{open-grid-2}
Tianshi Cao, Jingkang Wang, Yining Zhang, and Sivabalan Manivasagam.
\newblock Babyai++: Towards grounded-language learning beyond memorization.
\newblock {\em CoRR}, abs/2004.07200, 2020.

\bibitem{open-grid-1}
Maxime Chevalier-Boisvert, Dzmitry Bahdanau, Salem Lahlou, Lucas Willems,
  Chitwan Saharia, Thien~Huu Nguyen, and Yoshua Bengio.
\newblock Babyai: A platform to study the sample efficiency of grounded
  language learning.
\newblock {\em Learning}, 2018.

\bibitem{gcil}
Yiming Ding, Carlos Florensa, Pieter Abbeel, and Mariano Phielipp.
\newblock Goal-conditioned imitation learning.
\newblock {\em Advances in neural information processing systems}, 32, 2019.

\bibitem{dohare2021continual}
Shibhansh Dohare, A~Rupam Mahmood, and Richard~S Sutton.
\newblock Continual backprop: Stochastic gradient descent with persistent
  randomness.
\newblock {\em arXiv preprint arXiv:2108.06325}, 2021.

\bibitem{goexplore}
Adrien Ecoffet, Joost Huizinga, Joel Lehman, Kenneth~O Stanley, and Jeff Clune.
\newblock First return, then explore.
\newblock {\em Nature}, 590(7847):580--586, 2021.

\bibitem{starcraft3}
Islam Elnabarawy, Kristijana Arroyo, and Donald~C. Wunsch.
\newblock Starcraft ii build order optimization using deep reinforcement
  learning and monte-carlo tree search.
\newblock {\em arXiv: Learning}, 2020.

\bibitem{impala}
Lasse Espeholt, Hubert Soyer, Remi Munos, Karen Simonyan, Vlad Mnih, Tom Ward,
  Yotam Doron, Vlad Firoiu, Tim Harley, Iain Dunning, et~al.
\newblock Impala: Scalable distributed deep-rl with importance weighted
  actor-learner architectures.
\newblock In {\em ICML}, pages 1407--1416. PMLR, 2018.

\bibitem{open-indoor-4}
Linxi Fan, Guanzhi Wang, De-An Huang, Zhiding Yu, Li Fei-Fei, Yuke Zhu, and
  Animashree Anandkumar.
\newblock Secant: Self-expert cloning for zero-shot generalization of visual
  policies.
\newblock {\em arXiv: Learning}, 2021.

\bibitem{minedojo}
Linxi Fan, Guanzhi Wang, Yunfan Jiang, Ajay Mandlekar, Yuncong Yang, Haoyi Zhu,
  Andrew Tang, De-An Huang, Yuke Zhu, and Anima Anandkumar.
\newblock Minedojo: Building open-ended embodied agents with internet-scale
  knowledge.
\newblock {\em arXiv preprint arXiv:2206.08853}, 2022.

\bibitem{gcsl}
Dibya Ghosh, Abhishek Gupta, Ashwin Reddy, Justin Fu, Coline~Manon Devin,
  Benjamin Eysenbach, and Sergey Levine.
\newblock Learning to reach goals via iterated supervised learning.
\newblock In {\em International Conference on Learning Representations}, 2021.

\bibitem{minerl-rel-2}
William~H. Guss, Mario~Ynocente Castro, Sam Devlin, Brandon Houghton,
  Noboru~Sean Kuno, Crissman Loomis, Stephanie Milani, Sharada~P. Mohanty,
  Keisuke Nakata, Ruslan Salakhutdinov, John Schulman, Shinya Shiroshita,
  Nicholay Topin, Avinash Ummadisingu, and Oriol Vinyals.
\newblock The minerl 2020 competition on sample efficient reinforcement
  learning using human priors.
\newblock {\em arXiv: Learning}, 2021.

\bibitem{minerl-rel-1}
William~H Guss, Cayden Codel, Katja Hofmann, Brandon Houghton, Noboru Kuno,
  Stephanie Milani, Sharada Mohanty, Diego~Perez Liebana, Ruslan Salakhutdinov,
  Nicholay Topin, et~al.
\newblock Neurips 2019 competition: the minerl competition on sample efficient
  reinforcement learning using human priors.
\newblock {\em arXiv preprint arXiv:1904.10079}, 2019.

\bibitem{minerl}
William~H. Guss, Brandon Houghton, Nicholay Topin, Phillip Wang, Cayden Codel,
  Manuela Veloso, and Ruslan Salakhutdinov.
\newblock Minerl: A large-scale dataset of minecraft demonstrations.
\newblock {\em international joint conference on artificial intelligence},
  2019.

\bibitem{resnet}
Kaiming He, Xiangyu Zhang, Shaoqing Ren, and Jian Sun.
\newblock Deep residual learning for image recognition.
\newblock In {\em Proceedings of the IEEE conference on computer vision and
  pattern recognition}, pages 770--778, 2016.

\bibitem{film4}
Eric Jang, Alex Irpan, Mohi Khansari, Daniel Kappler, Frederik Ebert, Corey
  Lynch, Sergey Levine, and Chelsea Finn.
\newblock Bc-z: Zero-shot task generalization with robotic imitation learning.
\newblock In {\em Conference on Robot Learning}, pages 991--1002. PMLR, 2022.

\bibitem{bias3}
Menglin Jia, Luming Tang, Bor{-}Chun Chen, Claire Cardie, Serge~J. Belongie,
  Bharath Hariharan, and Ser{-}Nam Lim.
\newblock Visual prompt tuning.
\newblock In Shai Avidan, Gabriel~J. Brostow, Moustapha Ciss{\'{e}},
  Giovanni~Maria Farinella, and Tal Hassner, editors, {\em Computer Vision -
  {ECCV} 2022 - 17th European Conference, Tel Aviv, Israel, October 23-27,
  2022, Proceedings, Part {XXXIII}}, volume 13693 of {\em Lecture Notes in
  Computer Science}, pages 709--727. Springer, 2022.

\bibitem{malmo}
Matthew Johnson, Katja Hofmann, Tim~J. Hutton, and David~Michael Bignell.
\newblock The malmo platform for artificial intelligence experimentation.
\newblock {\em international joint conference on artificial intelligence},
  2016.

\bibitem{open-maze-2}
Arthur Juliani, Ahmed Khalifa, Vincent~Pierre Berges, Jonathan Harper, Ervin
  Teng, Hunter Henry, Adam Crespi, Julian Togelius, and Danny Lange.
\newblock Obstacle tower: A generalization challenge in vision, control, and
  planning.
\newblock {\em international joint conference on artificial intelligence},
  2019.

\bibitem{minerl-rel-3}
Anssi Kanervisto, Stephanie Milani, Karolis Ramanauskas, Nicholay Topin,
  Zichuan Lin, Junyou Li, Jianing Shi, Deheng Ye, Qiang Fu, Wei Yang, Weijun
  Hong, Zhongyue Huang, Haicheng Chen, Guangjun Zeng, Yue Lin, Vincent Micheli,
  Eloi Alonso, Fran\\c\{c\}ois Fleuret, Alexander Nikulin, Yury Belousov, Oleg
  Svidchenko, and Aleksei Shpilman.
\newblock Minerl diamond 2021 competition: Overview, results, and lessons
  learned.
\newblock {\em neural information processing systems}, 2022.

\bibitem{bilinear}
Jin-Hwa Kim, Kyoung-Woon On, Woosang Lim, Jeonghee Kim, Jung-Woo Ha, and
  Byoung-Tak Zhang.
\newblock Hadamard product for low-rank bilinear pooling.
\newblock {\em arXiv preprint arXiv:1610.04325}, 2016.

\bibitem{juewu}
Zichuan Lin, Junyou Li, Jianing Shi, Deheng Ye, Qiang Fu, and Wei Yang.
\newblock Juewu-mc: Playing minecraft with sample-efficient hierarchical
  reinforcement learning.
\newblock {\em arXiv preprint arXiv:2112.04907}, 2021.

\bibitem{progress1}
Chih{-}Yao Ma, Jiasen Lu, Zuxuan Wu, Ghassan AlRegib, Zsolt Kira, Richard
  Socher, and Caiming Xiong.
\newblock Self-monitoring navigation agent via auxiliary progress estimation.
\newblock In {\em 7th International Conference on Learning Representations,
  {ICLR} 2019, New Orleans, LA, USA, May 6-9, 2019}. OpenReview.net, 2019.

\bibitem{progress3}
Chih-Yao Ma, Zuxuan Wu, Ghassan AlRegib, Caiming Xiong, and Zsolt Kira.
\newblock The regretful agent: Heuristic-aided navigation through progress
  estimation.
\newblock In {\em Proceedings of the IEEE/CVF conference on Computer Vision and
  Pattern Recognition}, pages 6732--6740, 2019.

\bibitem{SEIHAI}
Hangyu Mao, Chao Wang, Xiaotian Hao, Yihuan Mao, Yiming Lu, Chengjie Wu, Jianye
  Hao, Dong Li, and Pingzhong Tang.
\newblock Seihai: A sample-efficient hierarchical ai for the minerl
  competition.
\newblock In {\em International Conference on Distributed Artificial
  Intelligence}, pages 38--51. Springer, 2021.

\bibitem{atari}
Volodymyr Mnih, Koray Kavukcuoglu, David Silver, Alex Graves, Ioannis
  Antonoglou, Daan Wierstra, and Martin Riedmiller.
\newblock Playing atari with deep reinforcement learning.
\newblock {\em arXiv: Learning}, 2013.

\bibitem{film3}
Junhyuk Oh, Satinder Singh, Honglak Lee, and Pushmeet Kohli.
\newblock Zero-shot task generalization with multi-task deep reinforcement
  learning.
\newblock In {\em International Conference on Machine Learning}, pages
  2661--2670. PMLR, 2017.

\bibitem{film1}
Ethan Perez, Florian Strub, Harm De~Vries, Vincent Dumoulin, and Aaron
  Courville.
\newblock Film: Visual reasoning with a general conditioning layer.
\newblock In {\em Proceedings of the AAAI Conference on Artificial
  Intelligence}, volume~32, 2018.

\bibitem{gato}
Scott Reed, Konrad Zolna, Emilio Parisotto, Sergio~Gomez Colmenarejo, Alexander
  Novikov, Gabriel Barth-Maron, Mai Gimenez, Yury Sulsky, Jackie Kay,
  Jost~Tobias Springenberg, et~al.
\newblock A generalist agent.
\newblock {\em arXiv preprint arXiv:2205.06175}, 2022.

\bibitem{habitat}
Manolis Savva, Abhishek Kadian, Oleksandr Maksymets, Yili Zhao, Erik Wijmans,
  Bhavana Jain, Julian Straub, Jia Liu, Vladlen Koltun, Jitendra Malik, Devi
  Parikh, and Dhruv Batra.
\newblock Habitat: A platform for embodied ai research.
\newblock {\em international conference on computer vision}, 2019.

\bibitem{udrl}
Juergen Schmidhuber.
\newblock Reinforcement learning upside down: Don't predict rewards--just map
  them to actions.
\newblock {\em arXiv preprint arXiv:1912.02875}, 2019.

\bibitem{open-indoor-2}
Bokui Shen, Fei Xia, Chengshu Li, Roberto Mart{\'i}n-Mart{\'i}n, Linxi Fan,
  Guanzhi Wang, Shyamal Buch, Claudia D'Arpino, Sanjana Srivastava, Lyne~P.
  Tchapmi, Micael Tchapmi, Kent Vainio, Li Fei-Fei, and Silvio Savarese.
\newblock igibson, a simulation environment for interactive tasks in large
  realistic scenes.
\newblock {\em intelligent robots and systems}, 2020.

\bibitem{go}
David Silver, Julian Schrittwieser, Karen Simonyan, Ioannis Antonoglou, Aja
  Huang, Arthur Guez, Thomas Hubert, Lucas Baker, Matthew Lai, Adrian Bolton,
  et~al.
\newblock Mastering the game of go without human knowledge.
\newblock {\em nature}, 550(7676):354--359, 2017.

\bibitem{open-indoor-3}
Sanjana Srivastava, Chengshu Li, Michael Lingelbach, Roberto
  Mart{\'i}n-Mart{\'i}n, Fei Xia, Kent Vainio, Zheng Lian, Cem Gokmen, Shyamal
  Buch, C.~Karen Liu, Silvio Savarese, Hyowon Gweon, Jiajun Wu, and Li Fei-Fei.
\newblock Behavior: Benchmark for everyday household activities in virtual,
  interactive, and ecological environments.
\newblock {\em Conference on Robot Learning}, 2021.

\bibitem{dm_control}
Yuval Tassa, Yotam Doron, Alistair Muldal, Tom Erez, Yazhe Li, Diego de~Las
  Casas, David Budden, Abbas Abdolmaleki, Josh Merel, Andrew Lefrancq, et~al.
\newblock Deepmind control suite.
\newblock {\em arXiv preprint arXiv:1801.00690}, 2018.

\bibitem{open-maze-1}
Open Ended~Learning Team, Adam Stooke, Anuj Mahajan, Catarina Barros, Charlie
  Deck, Jakob Bauer, Jakub Sygnowski, Maja Trebacz, Max Jaderberg,
  Micha{\"{e}}l Mathieu, Nat McAleese, Nathalie Bradley{-}Schmieg, Nathaniel
  Wong, Nicolas Porcel, Roberta Raileanu, Steph Hughes{-}Fitt, Valentin
  Dalibard, and Wojciech~Marian Czarnecki.
\newblock Open-ended learning leads to generally capable agents.
\newblock {\em CoRR}, abs/2107.12808, 2021.

\bibitem{mujoco}
Emanuel Todorov, Tom Erez, and Yuval Tassa.
\newblock Mujoco: A physics engine for model-based control.
\newblock {\em intelligent robots and systems}, 2012.

\bibitem{starcraft}
Oriol Vinyals, Igor Babuschkin, Wojciech~M Czarnecki, Micha{\"e}l Mathieu,
  Andrew Dudzik, Junyoung Chung, David~H Choi, Richard Powell, Timo Ewalds,
  Petko Georgiev, et~al.
\newblock Grandmaster level in starcraft ii using multi-agent reinforcement
  learning.
\newblock {\em Nature}, 575(7782):350--354, 2019.

\bibitem{starcraft2}
Oriol Vinyals, Timo Ewalds, Sergey Bartunov, Petko Georgiev, Alexander
  Vezhnevets, Michelle Yeo, Alireza Makhzani, Heinrich K{\"u}ttler, John~P.
  Agapiou, Julian Schrittwieser, John Quan, Stephen Gaffney, Stig Petersen,
  Karen Simonyan, Tom Schaul, Hado van Hasselt, David Silver, Timothy~P.
  Lillicrap, Kevin Calderone, Paul Keet, Anthony Brunasso, David Lawrence,
  Anders Ekermo, Jacob Repp, and Rodney Tsing.
\newblock Starcraft ii: A new challenge for reinforcement learning.
\newblock {\em arXiv: Learning}, 2017.

\bibitem{open-maze-3}
Rui Wang, Joel Lehman, Jeff Clune, and Kenneth~O. Stanley.
\newblock Paired open-ended trailblazer (poet): Endlessly generating
  increasingly complex and diverse learning environments and their solutions.
\newblock {\em arXiv: Neural and Evolutionary Computing}, 2019.

\bibitem{progress4}
Joel Ye, Dhruv Batra, Abhishek Das, and Erik Wijmans.
\newblock Auxiliary tasks and exploration enable objectnav.
\newblock {\em arXiv preprint arXiv:2104.04112}, 2021.

\bibitem{meta-world}
Tianhe Yu, Deirdre Quillen, Zhanpeng He, Ryan Julian, Karol Hausman, Chelsea
  Finn, and Sergey Levine.
\newblock Meta-world: A benchmark and evaluation for multi-task and meta
  reinforcement learning.
\newblock 2019.

\bibitem{metaworld}
Tianhe Yu, Deirdre Quillen, Zhanpeng He, Ryan Julian, Karol Hausman, Chelsea
  Finn, and Sergey Levine.
\newblock Meta-world: A benchmark and evaluation for multi-task and meta
  reinforcement learning.
\newblock In {\em Conference on robot learning}, pages 1094--1100. PMLR, 2020.

\bibitem{bias2}
Elad~Ben Zaken, Shauli Ravfogel, and Yoav Goldberg.
\newblock Bitfit: Simple parameter-efficient fine-tuning for transformer-based
  masked language-models.
\newblock {\em CoRR}, abs/2106.10199, 2021.

\end{thebibliography}
}

\clearpage
\appendix

\section{Experimental Details}

\subsection{Observation and Action Space} \label{sec:space}
The agent receives identical information as human players do. The observation space primarily comprises four components: 1) ego-centric RGB frames, 2) voxels (surrounding blocks), 3) GPS locations (the agent's three-dimensional coordinates), and 4) compass (pitch/yaw angles). These are shaped as $(3, 480, 640)$, $(3, 3, 3)$, $(3,)$, and $(2, )$, respectively. 
It is important to note that the agent \textbf{does not know} the precise location of the target object. Instead, the agent can only obtain information about the target object by examining the pixel image. 
The RGB frames are resized to a shape of $(3, 128, 128)$ using bilinear interpolation before being fed into the networks. 
At each step, the agent must execute a movement action, camera action, and functional action. A compound action space is employed, consisting of a multi-discrete space with six dimensions: 1) forward and backward, 2) move left and right, 3) jump, sneak and sprint, 4) camera delta pitch, 5) camera delta yaw, and 6) functional actions (attack and use). 
The original delta camera degree, which ranges from -180 to 180, is discretized into 11 bins. 
As this paper's primary focus is on resource collection rather than item crafting, actions related to crafting are omitted. 


\subsection{Data Collection Pipeline} \label{sec:data_collection_pipeline}
Our data collection pipeline collects high-quality goal-conditioned demonstrations with actions. 
The core idea is to train a proxy policy with non-goal demonstrations and roll out in customized environments, then filter the demonstrations according to the achievement. 
Generally, the pipeline consists of six steps: 1) collect online videos, 2) clean and label the videos, 3) train a proxy policy, 4) customize the environments, 5) roll out the proxy policy, and 6) filter by the accomplishments. 

Video-Pretraining\!~\cite{VPT} is ideally suited for stages 1-3. It begins by amassing a vast dataset of Minecraft videos, sourced from the web using relevant keywords. Given that collected videos often feature overlaid artifacts, the process filters out videos without visual artifacts and those from survival mode. Next, an Inverse Dynamics Model (IDM) is trained to label these videos with actions, yielding demonstrations for proxy policy training. We directly employ the pretrained VPT\!~\cite{VPT} as our proxy policy. 
In stage 4, we utilize APIs supplied by MineDojo\!~\cite{minedojo} to create environments tailored to each task's success criteria. During stage 5, we deploy the proxy policy, recording successful trajectories and their corresponding achieved goals. The environment is reset once the episode concludes or the goal is accomplished, ensuring trajectory independence.

Notably, we execute the proxy policy rollout in parallel using 16 processes on 4 A40 GPUs, generating 0.5GB of demonstrations per minute (without leveraging video compression algorithm during storing frames). This approach minimizes human intervention and enhances data collection efficiency. In total, we have gathered 215GB, 289GB, and 446GB of goal-conditioned demonstrations from \textbf{\texttt{Plains}}, \textbf{\texttt{Flat}}, and \textbf{\texttt{Forest}} environments, respectively. 




\subsection{Implementation} \label{sec:implementation}

\noindent \textbf{Horizon discretization. }
As the horizon illustrates the number of steps required to attain the desired objective, it is infeasible to precisely determine the exact value. In practice, we suggest dividing the original horizon into 16 distinct segments: $\left[0, 10\right) \rightarrow 0$, $\left[10, 20\right) \rightarrow 1$, $\left[20, 30\right) \rightarrow 2$, $\cdots$, $\left[90, 100\right) \rightarrow 9$, $\left[100, 120\right) \rightarrow 10$, $\left[120, 140\right) \rightarrow 11$, $\cdots$, $\left[180, 200\right) \rightarrow 14$, and $\left[200, \infty\right) \rightarrow 15$. 
In this approach, each segment inherently represents a phase that signifies the level of task completion. Consequently, the horizon prediction issue can be framed as a multi-class problem. It is important to note that the method of discretization is not singular and merits further exploration in the future.

\noindent \textbf{Training.} 
The observation of RGB image is scaled into $128\times 128$ where no data augmentation is adopted. 
We train the policy with the AdamW optimizer and a linear learning rate decay. We use an initial learning rate of 0.0001, a batch size of 32, and a weight decay of 0.0001. 
Besides, we also use a warmup trick that the learning rate linearly increases from 0 to 0.0001 in 10k iterations. 
The policy is trained for 500k iterations on our collected dataset. It takes one day on a single A40 GPU. 
To train the baseline policies BC~({\footnotesize VPT}) and BC~({\footnotesize CLIP}), we only finetune the bias terms of their backbones, which is widely adopted by 
previous works\cite{bias1, bias2, bias3}. Also note that, to keep the architecture comparable, we only transfer model and weights of the backbone from vpt model and MineCLIP model while replace their transformer architecture with ours. 

\noindent \textbf{Evaluation.}
During the evaluation, the maximum episode length is empirically set to 600, 600, and 300 for the \textbf{\texttt{Flat}}, \textbf{\texttt{Plains}}, and \textbf{\texttt{Forest}} biomes, respectively. In most instances, the agent is able to complete the assigned tasks within these limits.
Furthermore, in our adaptive horizon prediction module, the hyperparameter $c$ is empirically set to $3$. The model is evaluated every 10,000 gradient updates. During each evaluation round, each goal is assessed 10 times to compute the \textit{Success Rate} and \textit{Precision} metrics.
For the ablation study, we utilize the checkpoint after 500,000 training iterations, evaluate each goal 200 times, and report the average metrics in Table \ref{tab:cross_biome} and Figure \ref{fig:ablation}.

\section{Horizon Distribution Analysis} \label{sec:horizon_distribuition}
\begin{figure*}[t]
    \centering
    \includegraphics[scale = 0.52]{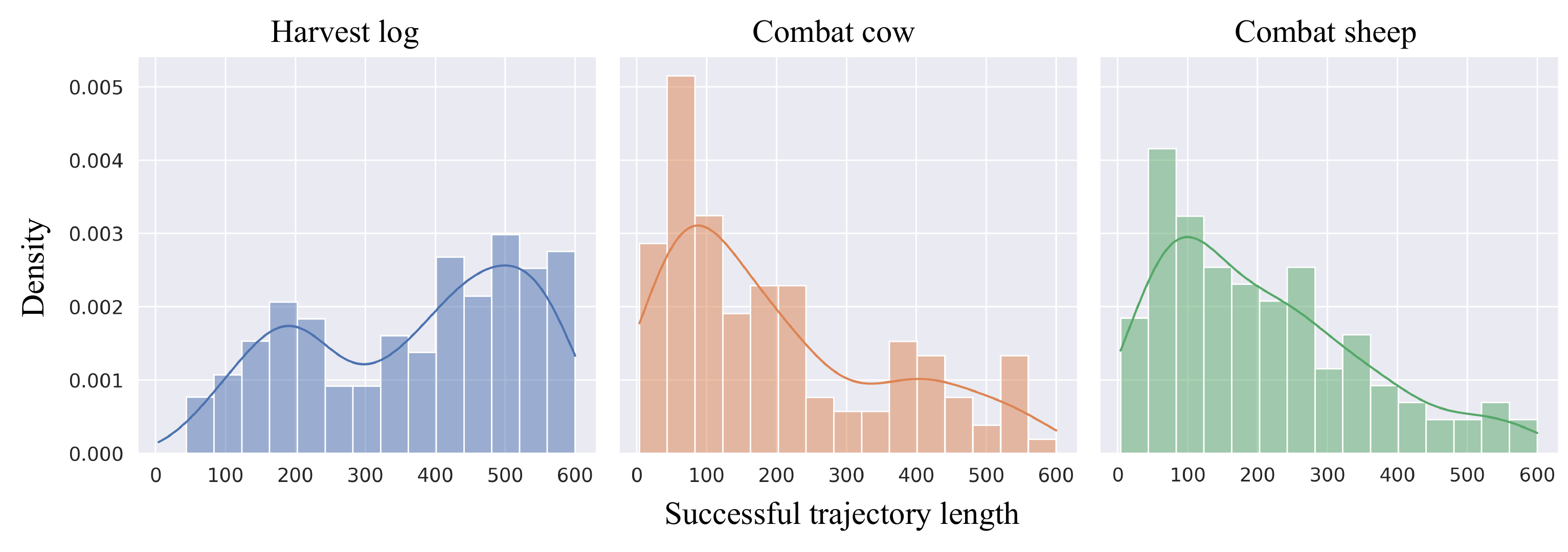}
    
    \caption{
        \textbf{Successful trajectory distribution of different tasks in open-ended Minecraft}. The distribution is long-tailed, making it hard to learn goal-conditioned policies with a fixed horizon.
    }
    \label{fig:traj_dis}
\end{figure*}
To further emphasize the importance of our adaptive horizon prediction module, we have visualized the distribution of successful trajectory lengths for various tasks in Minecraft, as shown in Figure~\ref{fig:traj_dis}. These successful trajectories were gathered from agents trained using single-task behavior cloning (with a randomly initialized Impala CNN as the backbone) in the \textbf{\texttt{Plains}} biome.

As depicted in Figure~\ref{fig:traj_dis}, the distribution of successful trajectory lengths in the open-world setting exhibits a long tail, making it challenging to train a policy with a fixed horizon. This can be attributed to Minecraft's extensive explorable space, partial observation properties, and non-stationary dynamics, which set it apart from other popular multi-task, closed-ended environments like Meta-World \cite{metaworld}.

Consequently, the minimum number of steps needed for an agent to achieve its goal varies across different environments and episodes. The episode length typically hinges on the relative position and terrain constraints between the target object and the agent's initial position. An added layer of complexity arises when no target objects are near the agent's starting location, necessitating large-scale exploration (i.e., a larger horizon). Once the agent locates the target object, it must track it until the relevant skill can be executed on the object (e.g., killing or harvesting). This demands that the agent remain aware of its current stage.

Our proposed adaptive horizon prediction module incorporates the horizon as an additional condition for the policy. The policy explicitly takes into account the remaining time steps needed to achieve specific goals. Our experiments in Section~\ref{sec:ablation_horizon} demonstrate that the adaptive horizon prediction module and the horizon loss $\mathcal{L}_{h}$ effectively enhance the success rate in open-world environments with such distributions.

\section{Limitation and Future Work}
In essence, our approach hinges on trajectories labeled with goals, which enables it to generalize across various domains, provided that such data is accessible. When only video segments labeled with actions are available, we can employ a goal predictor to assign goal labels to these clips. This can also be achieved by utilizing zero-shot models, such as CLIP. Moreover, if action labels are absent in these clips, we can resort to training an inverse dynamics model, as demonstrated in VPT. Undoubtedly, these present intriguing avenues for future exploration

\end{document}